\documentclass[preprint,12pt,authoryear]{elsarticle}

\usepackage{amssymb}

\journal{a journal}

\begin{document}

\begin{frontmatter}



\title{A new explainable DTM generation algorithm with airborne LIDAR data: grounds are smoothly connected eventually}


\author[inst1]{Hunsoo Song}

\affiliation[inst1]{organization={Lyles School of Civil Engineering},
            addressline={Purdue University}, 
            city={West Lafayette},
            postcode={47907}, 
            state={Indiana},
            country={USA}}

\author[inst1]{Jinha Jung}


\begin{abstract}
The digital terrain model (DTM) is fundamental geospatial data for various studies in urban, environmental, and Earth science. The reliability of the results obtained from such studies can be considerably affected by the errors and uncertainties of the underlying DTM. Numerous algorithms have been developed to mitigate the errors and uncertainties of DTM. However, most algorithms involve tricky parameter selection and complicated procedures that make the algorithm’s decision rule obscure, so it is often difficult to explain and predict the errors and uncertainties of the resulting DTM. Also, previous algorithms often consider the local neighborhood of each point for distinguishing non-ground objects, which limits both search radius and contextual understanding and can be susceptible to errors particularly if point density varies. This study presents an open-source DTM generation algorithm for airborne LiDAR data that can consider beyond the local neighborhood and whose results are easily explainable, predictable, and reliable. The key assumption of the algorithm is that grounds are smoothly connected while non-grounds are surrounded by areas having sharp elevation changes. The robustness and uniqueness of the proposed algorithm were evaluated in geographically complex environments through tiling evaluation compared to other state-of-the-art algorithms.
\end{abstract}



\begin{keyword}
Digital terrain model \sep Airborne laser scanning \sep LiDAR \sep Ground filtering
\end{keyword}

\end{frontmatter}


\section{Introduction}
\label{sec:Introduction}


The digital terrain model (DTM), also often referred to as the digital elevation model (DEM), is a 3-dimensional representation of the bare earth surface excluding any ground-standing objects like trees and buildings. DTM is an essential geospatial data for various studies, namely, hydrological modeling \citep{callow2007does,chaney2018harnessing,jarihani2015satellite}, glacier monitoring \citep{shean2019ice}, landslide monitoring \citep{jaboyedoff2012use,tseng2013application,kim2015pre}, land-cover classification \citep{rodriguez2012assessment,yan2015urban}, building mapping \citep{song2022towards}, forestry \citep{oh2022high,simpson2017assessment}, and agricultural management \citep{tarolli2020agriculture}. Since errors and uncertainties in DTM can significantly affect the knowledge gained from such studies, producing an accurate and reliable estimate of the terrain is crucial \citep{goulden2016sensitivity,wechsler2007uncertainties}. Generating DTM requires classifying the bare earth surface among all 3-dimensional coordinate measurements over the earth. Light detection and ranging (LiDAR) \citep{chen2017state}, radar \citep{farr2007shuttle, rizzoli2017generation}, and photogrammetry technologies \citep{turner2012automated,bhushan2021automated,shean2016automated} that can retrieve the 3-dimensional coordinates of the earth are generally used for producing DTM. Among different sources for producing DTM, airborne LiDAR data (airborne laser scanning) has become the most powerful sensor for generating high-resolution DTM in terms of its accuracy, and numerous algorithms have been developed for DTM generation. However, generating an accurate and reliable DTM with a scalable method remains a challenge. 

DTM generating algorithm with airborne LiDAR data usually necessitates the procedure of classifying ground and non-ground objects. In most cases, the available data for this binary classification is coordinate measurements of the earth’s surface. Therefore, geometrical shapes and associations among coordinates are used for the classification. Typically, DTM generating algorithms aim to make a decision based on the assumption that ground is generally smooth while non-ground objects have protruding shapes \citep{meng2010ground,chen2017state}. Considering that most DTM generating algorithms share the common goal of discriminating between smooth and protruding shapes, algorithms can be classified according to how to represent 3-dimensional coordinate measurements. Point cloud \citep{bartels2010threshold,bartels2006dtm,sithole2005filtering,vosselman2000slope,zhang2016easy}, triangulated irregular network (TIN) \citep{axelsson2000generation,sohn2002terrain,zhang2013filtering}, and image grid \citep{amirkolaee2022dtm,chen2012upward,gevaert2018deep,hu2016deep,lohmann2000approaches,mongus2013computationally,wack2002digital,zhang2003progressive} are the three most common representations of 3-dimensional coordinate measurements. 

First, point clouds-based algorithms typically consider each point’s relative coordinates with respect to its local neighboring points \citep{meng2010ground}. Then, the common way to classify non-ground points is based on a discriminant function describing slopes among a set of points. The point clouds representation has an advantage in that it can preserve and directly use the raw measurements, and it allows more flexible operations as its representation is non-gridded. However, handling outlier is difficult, and it often fails to produce reliable results particularly when the local point density is varying. Second, TIN represents 3-dimensional coordinate measurements as a continuous surface consisting of triangular facets, also referred to as a triangle mesh. The TIN representation allows algorithms to effectively use the local structure of point coordinates as each triangular facet can be assumed as an approximation of the local surface. However, it has the same disadvantages as point cloud representations in that they are irregularly spaced data that are difficult to process. Lastly, the image grid representation projects the point cloud into a 2-dimensional image grid and considers the elevation (Z) of coordinates as the pixel value of the image. In other words, this method creates a digital surface model (DSM) first and generates DTM. As it transforms 3-dimensional measurements into gridded data, it has advantages in that it allows morphological operation and the operation is conceptually simple. However, it necessarily distorts and compromises the original data. 

Regardless of which representation method is adopted, DTM generating algorithms try to classify ground from non-ground based on the assumption that ground is generally smooth while non-ground objects have protruding shapes. The difficult thing in the classification is that there is no clear boundary between “smoothness” and "protrudeness”. Non-ground objects also can have smooth surfaces, and the challenge is how large an area should be taken into account when classifying objects \citep{meng2010ground}. For example, a large building with a flat roof can be classified as non-ground only if the algorithm considers a larger area than the building. Otherwise, points near the center of the flat roof will be classified as ground. Yet, simply expanding the area of consideration does not help the problem. The more the algorithm considers a large area, the more variables there are, and the more the algorithm has to find a complex and sophisticated decision boundary. This often results in requiring a lot of parameter tuning for the algorithm, and in turn, the generalization capability of the algorithm would be degraded. In a nutshell, resolving uncertainties in measuring the smoothness and in determining area-to-consider is the key to the algorithm.

To be specific, DTM generating algorithms based on either point clouds or TIN generally set search radius and compute slopes or angles to adjacent points within the search radius for quantifying the smoothness. Then, the discriminant function to filter non-ground objects is the function of the search radius and slopes. Again, the challenging part is setting a proper size of the search radius. As object size varies considerably, the pre-determined discriminant function to filter non-ground objects is hard to generalize for diverse landscapes. Especially as point clouds and TIN representations operate with irregularly spaced data, defining suitable parameters for search radius and slope threshold can be more difficult. 

Even with the image grid representation that can take advantage of simple morphological operations easily, the discriminant function needs to determine a certain window (or kernel) size, conceptually the same as the search radius. Indeed, previous studies had difficulties in selecting the proper window size for the morphological operation as the shape and the size of various objects are hard to generalize \citep{lohmann2000approaches}. Also, setting a proper search radius often requires prior knowledge of the given area. To resolve the problem of pre-defined search radius, several algorithms have been developed to adaptively change their search radiuses \citep{zhang2003progressive}. However, as algorithms become more complex, they tend to require more computations and a larger number of parameters to produce satisfactory results, while losing generalization capability, becoming more difficult to set proper parameters, and making the resulting DTM difficult to predictable. 

Alternative recent methods for generating DTM algorithms include deep learning-based methods \citep{amirkolaee2022dtm,gevaert2018deep,hu2016deep} and a cloth simulation-based method \citep{zhang2016easy}. Deep learning-based algorithms usually regard DSM as an image and try to extract non-ground pixels similar to the common approach for computer-vision tasks, namely, semantic segmentation or object detection. Although deep learning-based methods produced promising results, they require a large number of labeled training samples and huge computation resources. Also, the quality of output is bounded by not only the LiDAR data but also the reference DTM for the training, and the trained model may not reproduce satisfactory results when the target area has different properties from that of the trained area. Unlike deep learning-based methods, the cloth simulation-based method assumes that a virtual cloth covering on the upside-down DSM could be a DTM. With this simulation of the physical process, the cloth simulation-based method also produced reasonable results and reduced the number of parameters to tune compared to those of the conventional methods. However, it still requires tricky parameter tunings and can cause errors when dealing with very large low buildings and terrains having unique shapes, such as bridges \citep{vstroner2021vegetation,yu2022unsupervised}. 

This paper presents a novel, open-source DTM generating algorithm for airborne LiDAR data based on the image grid representation. The algorithm projects point clouds into a finely gridded DSM so that DSM sufficiently preserves the information of the original point cloud. With the finely rasterized DSM, the algorithm uses a Sobel operator to calculate the slope information and classifies ground and non-ground based on a simple but novel assumption. The assumption is that any non-ground object is surrounded by a certain steep level of a slope while grounds are smoothly connected to each other eventually. Different from previous algorithms relying on the local neighborhood for defining a discriminant function, the proposed algorithm can consider beyond the local neighborhood and classifies non-ground objects based on their context. More importantly, as it classifies ground and non-ground based on a physically straightforward rule, slope, the parameter tuning is very easy and straightforward, and the results are explainable and predictable. The algorithm turns out to be robust in diverse scenarios, computationally efficient, and easy-to-use as it requires only a few parameters that can be easily determined by the user’s objective. In addition, the algorithm includes a feature that can detect and map the elevation of the water body. Therefore, users of this algorithm can expect a seamless, full, rasterized DTM over the entire area of interest with only raw data that includes XYZ coordinates of point observations. The algorithm will be publicly available via GitHub.

The remainder of this paper is organized as follows. Section 2 elaborates on the proposed DTM generating algorithm. Section 3 discusses experimental results in comparison with widely adopted DTM generating methods (i.e., cloth simulation filtering \citep{zhang2016easy} and TIN-based method \citep{axelsson2000generation}) and provides suggestions for parameter tuning. Section 4 concludes the paper.

\section{Proposed DTM generating algorithm}
\label{sec:Proposed DTM generating algorithm}

The proposed algorithm consists of the following main four steps: (1) finely rasterized DSM generation, (2) break-line mapping with the Sobel operator, (3) filtering non-ground objects, and (4) water mapping. The following subsection describes each of the four main steps in detail and provides a summary of the proposed algorithm.

\subsection{Finely rasterized DSM generation}
\label{sec:Finely rasterized DSM generation}

3-dimensional coordinates of the earth's surface are usually collected as point cloud data, and the point cloud is assumed to be proper enough to model DTM. However, the point cloud is still a set of observations of real-world entities, so it necessarily has a limitation in perfectly representing the world. In particular, sensors most widely used for DTM generation, including LiDAR, have limitations in that their outputs are irregularly spaced 3-dimensional coordinates and their local point densities are inevitably varying. Even if the same laser pulse rate has been used during the flight mission, the point spacing will inevitably be different as it is affected by many factors such as flight configuration, flight condition, and objects on the ground \citep{habib2011geometric,morsdorf2008assessment,yu2004effects}. This is the main reason why most algorithms based on point clouds or TIN representations require lots of parameters to tune and are hard to generalize in universal scenarios.

In contrast, the image grid representation provides regularly spaced data. Specifically, when point clouds are the observation of the earth's surface from airborne laser scanning, the image grid representation of point clouds is DSM. To generate DSM, users need to determine the grid size for rasterization, and the grid size often becomes the resolution of DTM. Depending on the grid size, multiple point clouds can share the same grid and some grids might not have any points. The down-sampling, which generates a large coarse grid DSM, often alongs with the rasterization to prevent void grids \citep{hyyppa2001segmentation,maltezos2018building,oh2022high}. Here, this type of rasterization with down-sampling is referred to as a “coarse rasterization”. The coarse rasterization can prevent void grids but results in data loss. To alleviate the data loss problem, this study adopts a “fine rasterization” that projects point clouds into a finely and regularly spaced image grid.

\begin{figure}
	\centering
	\includegraphics[width=1\textwidth]{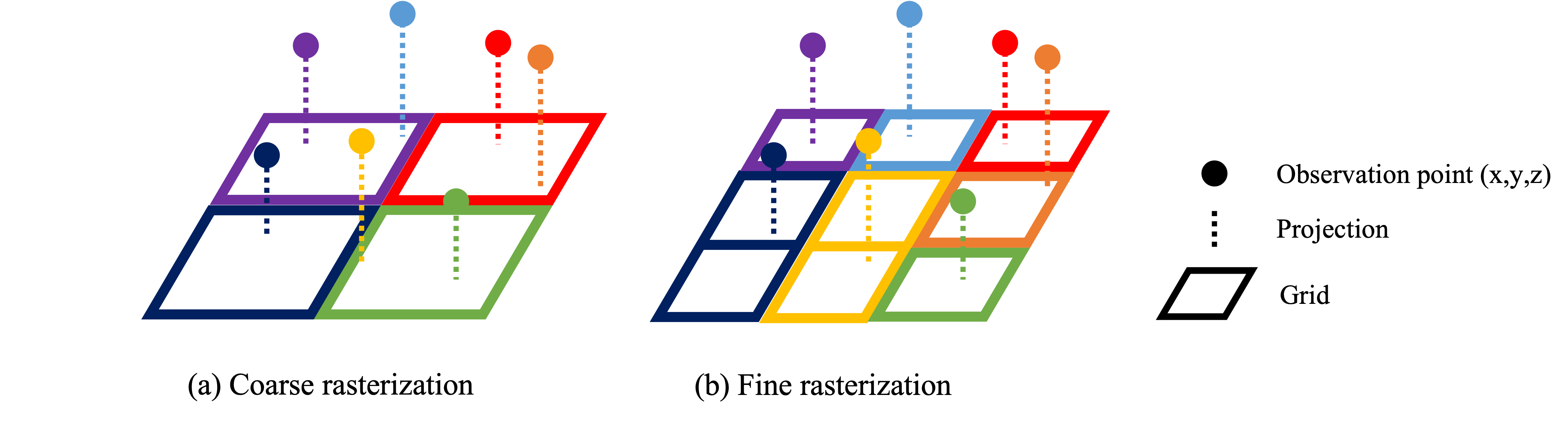}
	\caption{A comparison between (a) coarse rasterization and (b) fine rasterization. Fine rasterization is a relative concept compared to coarse rasterization. Either all or the majority of original point clouds can be preserved with a marginal horizontal displacement in the fine rasterization depending on the user-defined grid size. Different colors of the grid represent different elevations}
	\label{fig:fig1}
\end{figure}

Figure \ref{fig:fig1} provides a graphical illustration comparing coarse rasterization and fine rasterization. To project 7 observation points into an image grid, the coarse rasterization uses an image of 2 by 2 grids while the fine rasterization uses an image of 3 by 3 grids. The coarse rasterization does not have void grids while the fine rasterization has void grids initially. Void grids of the up-sampled image grid in the fine rasterization are filled up with the nearest neighbor interpolation. Fine rasterization can preserve more observation points with smaller displacement (registration) error than coarse rasterization, and thus it can help to generate precise, high-resolution DTM. 
However, the image grid representation either with coarse or fine rasterization may still result in data loss if some points coexist on the same grid. When multiple points occupy the same pixel, our DTM generation algorithm uses the lowest elevation point (the last return of LiDAR points) for the DSM value as the ground typically be the lowest elevation among neighboring points.


\subsection{Break-line mapping with the Sobel operator}
\label{sec:Break-line mapping with the Sobel operator}

\begin{figure}
	\centering
	\includegraphics[width=1\textwidth]{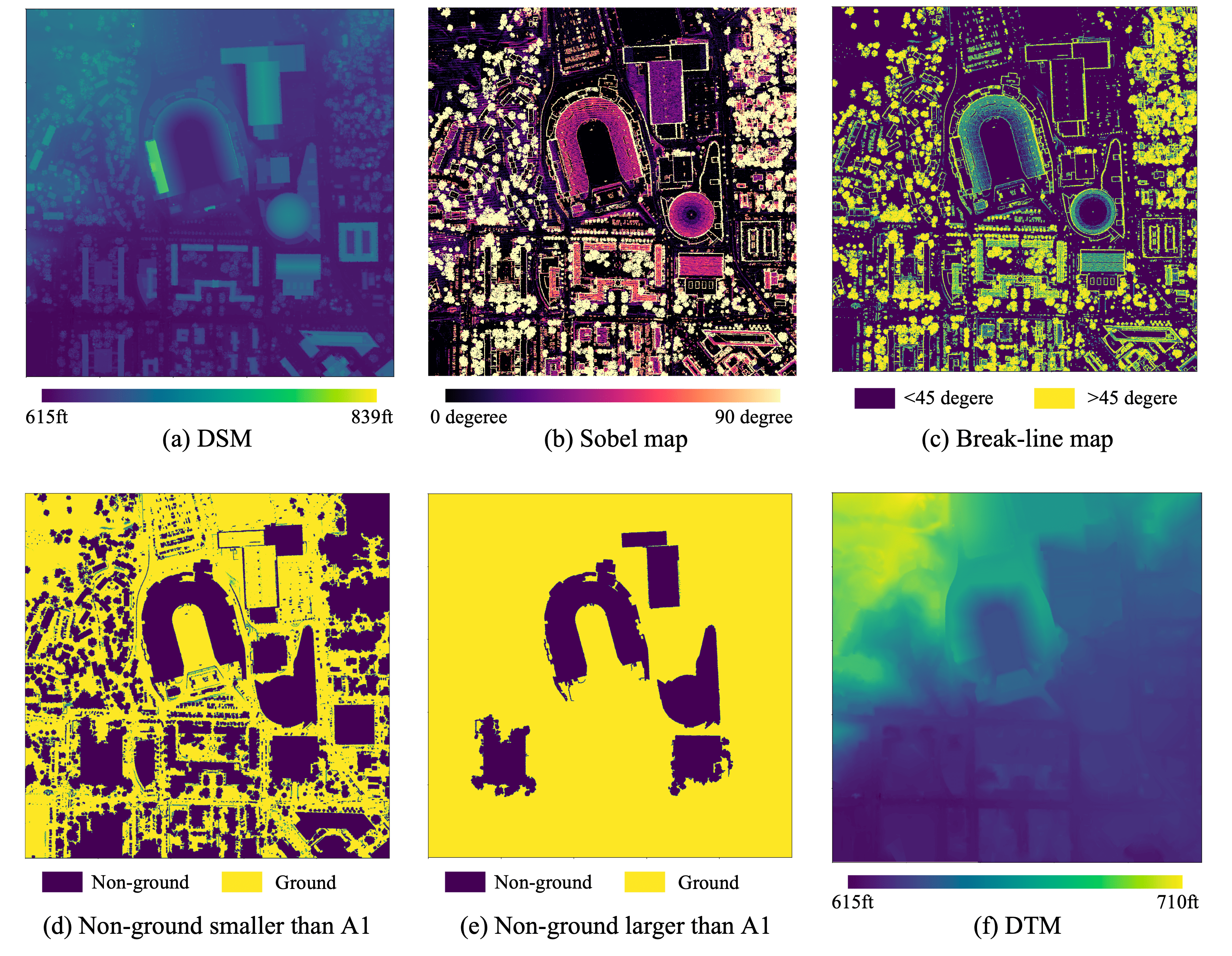}
	\caption{A procedure of the proposed DTM generation method: break-line mapping (a-c), non-ground filtering (d-e), and DTM (f)}
	\label{fig:fig2}
\end{figure}

A Sobel operator is a widely used kernel, particularly for edge detection in image processing applications as it essentially computes the gradient of the image intensity \citep{abdou1979quantitative}. Typically, a Sobel operator convolves two 3 by 3 kernels with the original image where kernels calculate derivatives of horizontal and vertical directions, respectively. Then, the Euclidean norm of two derivatives can describe the gradient of each pixel of the image. When the image is DSM, the gradient can be used to approximate the slope of the surface \citep{gelbman1984digital}. Therefore, DSM can be transformed into a slope map that describes the slope of the topography. Based on the slope map, we delineate a break-line map that shows the line where the topography is steeper than a certain level of slope. Figure \ref{fig:fig2} (a-c) illustrates an exemplary procedure of break-line mapping with the Sobel operator.

\subsection{Filtering non-ground objects}
\label{sec:Filtering non-ground objects}
A common assumption for DTM generation is that ground is generally smooth while non-ground objects have protruding shapes. In addition to this conventional assumption, we add the assumption that any non-ground object is surrounded by steep slopes while grounds are smoothly connected eventually. Thus, the proposed algorithm filters out any area surrounded by more than a certain degree of slope (i.e., the slope threshold). This assumption is reasonable and robust as hills, high-relief terrains, cliffs, mountain ranges, valleys, and overpasses are eventually connected to smooth surfaces in most cases; while non-ground objects like buildings and trees are enclosed by steep slopes or break-lines. In addition, since the break-line can span infinitely, the algorithm is not confined to the local neighborhood but can consider the global neighborhood and can classify non-ground objects regardless of their sizes and shapes. As a result, a non-ground object in our algorithm is clearly defined as a set of break-lines and an area surrounded by break-lines. Lastly, pixels classified as non-ground objects are masked and linearly interpolated with neighboring ground elevations to produce a seamless rasterized DTM.

The essential parameter of the proposed algorithm is the slope threshold parameter which is in charge of delineating break-lines. We claim that the parameter clearly possesses a physically meaningful value, and it can be easily tuned based on the user’s objective and topographical characteristics. We set the slope threshold of 45 degrees as default because it is robust enough to produce reliable DTM in most topographies. The impact and suggestions for parameter selection are discussed in Section 3.2.1.

\begin{figure}
	\centering
	\includegraphics[width=1\textwidth]{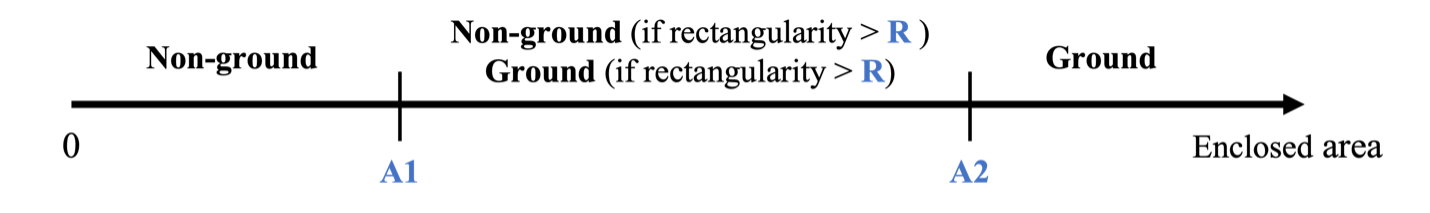}
	\caption{A ground and non-ground classification rule of the proposed DTM generation method}
	\label{fig:fig3}
\end{figure}

A few additional considerations were put into the algorithm to increase its scalability and facilitate its practical usage. First, some non-ground objects can lie on the edge of any given DSM layer. In this case, the non-ground object cannot be classified as a non-ground object because it is not fully surrounded by the break-line but is partially opened due to the limited data extent (the extent of given DSM layer). Therefore, the algorithm set the edge of the DSM as a break-line initially. Note that this action will enclose the ground that was originally not enclosed with break-lines. To prevent this issue, a decision based on the area was made to determine whether an enclosed area is ground or non-ground: First, if the enclosed area is smaller than a low limit (A1), the enclosed area is determined as a non-ground. Second, if the enclosed area is larger than a high limit (A2), it is determined as ground. Third, for the enclosed area between a low limit (A1) and a high limit (A2), a metric called the rectangularity that describes the ratio of the enclosed area to its minimum bounding rectangle area was considered. If the rectangularity is larger than a certain value (R), the enclosed area is classified as a non-ground, otherwise, it is classified as ground. This decision is based on the assumption that relatively large non-ground objects are mostly large buildings with rectangular shapes. Also, as all areas are eventually bounded by the size of the data extent, A2 is necessary. We set A1, A2, and R as 40,000$m^2$, 100,000$m^2$, and 50\% as a default. This is because objects larger than 40,000$m^2$ are rectangular-shaped buildings in most cases. We found these default values hardly produce artifacts. Figure \ref{fig:fig2} (d-e) displays an exemplary procedure of non-ground filtering, and Figure \ref{fig:fig2} (f) shows the final DTM. Figure \ref{fig:fig3} illustrates a ground and non-ground classification rule of the proposed DTM generation method.

\subsection{Water mapping}
\label{sec:Water mapping}
As the main task of DTM generation is typically considered as a classification of ground and non-ground, a group of algorithms called the “ground filtering algorithm” has received lots of attention instead of DTM generation \citep{meng2010ground}. Also, studies often evaluate the performance of DTM generation based on several binary classification metrics \citep{sithole2004experimental,mongus2013computationally,hu2015semi}. However, the ground filtering algorithm is usually limited in its purpose for classifying ground and non-ground by its definition and is not for mapping a full extent of a digital map that include both grounds and water bodies. Also, in general, external sources for water mapping are readily available due to lots of accumulated remotely sensed imagery \citep{huang2018detecting}. Perhaps these are the reasons why most DTM generation algorithms have ignored water body mapping \citep{hu2016deep,gevaert2018deep,amirkolaee2022dtm} even if subsequent analyses based on DTM often require a water map. However, the use of external data sources can lead to errors due to registration issues or mismatches in spatial and temporal resolution with the LiDAR data. Also, obscuration from clouds can be a problem when timely mapping is needed. In addition, a water map itself can be helpful to prevent errors in DTM mapping \citep{susaki2012adaptive}.

Therefore, we include a function that can extract water bodies and their elevations as well into our open-source workflow of DTM generation. In the proposed algorithm, water pixels are identified based on the assumption that the point density over water bodies is much lower than in non-water areas as water bodies hardly reflect laser points. With the finely rasterized DSM before the interpolation, the average point density (P) of a given scan is calculated by dividing the number of non-void grids by the number of total grids. Then, the number of non-void grids in a certain size of a sliding window (N pixels) will have a binomial distribution B(N, P). Specifically, B(N, P/2) was used to compensate for the imbalance of point density due to scanning overlap and to avoid overly detecting water. Based on the binomial distribution, lower confidence bound was used for the decision boundary of water classification. We used a window of 9 by 9 and a confidence level of 4 as a default. Lastly, the elevation of water bodies was selected by taking the 10th percentile of elevations among each water segment to prevent outliers. These parameters were set empirically and found to be robust in diverse topographic airborne laser scanning, but it is worth noting that elevation values cannot guarantee the true elevation as observations of water bodies contain lots of noise. This is because a LiDAR for topographic mapping commonly uses a near-infrared laser, which is absorbed by water and cannot reflect the laser point. Moreover, the elevation of the water bodies is dynamic in nature due to the water cycle. A more detailed description and impacts of water-related parameters are provided in Section 3.2.3. 

\subsection{Summary of DTM generation algorithm}
\label{sec:Summary of DTM generation algorithm}

The proposed DTM generation algorithm can convert a points cloud of airborne laser scanning to a rasterized DTM. The algorithm starts by generating the finely rasterized DSM to keep original points and transform the data regularly gridded. Based on the assumption that all non-ground objects are enclosed by a certain level of a steep slope, the algorithm delineates a break-line map with a Sobel operator and classifies non-ground objects based on the rectangularity and the size of the enclosed area. Finally, water mapping is performed considering the point density. Our method performs in an end-to-end manner and is easy to use as the meanings of parameters are very straightforward. Also, its results and errors are explainable and predictable in general, which can greatly reduce uncertainties in the resulting DTM.

\section{Experiments}
\label{sec:Experiments}

\begin{figure}
	\centering
	\includegraphics[width=1\textwidth]{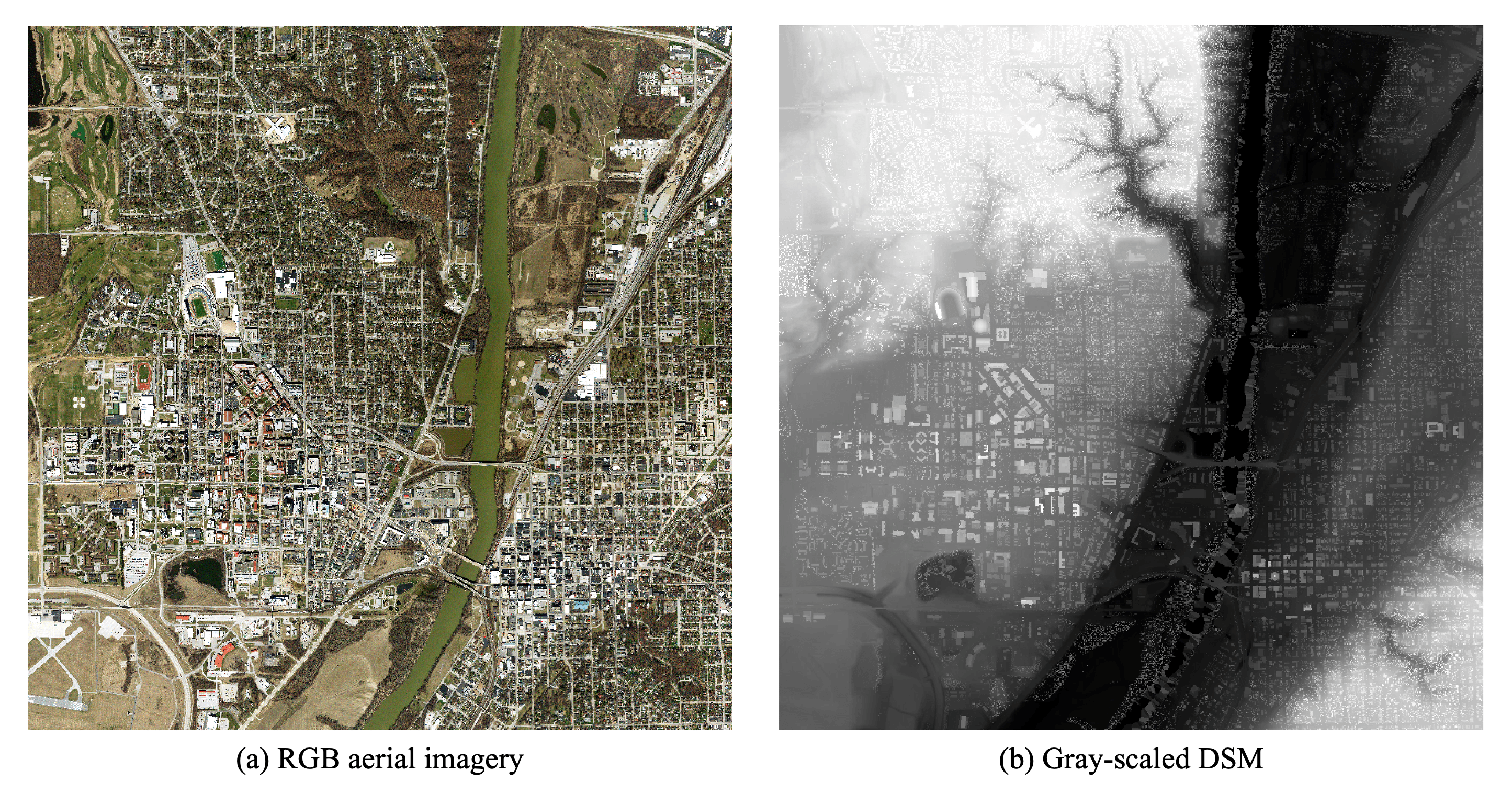}
	\caption{RGB aerial imagery and gray-scaled DSM over Purdue University Dataset}
	\label{fig:fig4}
\end{figure}

The proposed DTM generation method (“OUR”) was compared with two of the most popular DTM generation methods, the cloth simulation-based method (“CSF”) \citep{zhang2016easy} and TIN-based ground filtering method \citep{axelsson2000generation} implemented in LAStools (“LAS”) \footnote{https://rapidlasso.com/lastools/}. To compare their performances, an experimental area consisting of diverse landscapes, such as buildings, hilly forests, cropland, river, and deep valleys, was selected. We will refer to the study area as “Purdue University Dataset” hereafter. Purdue University Dataset includes West Lafayette and Lafayette, Indiana, United States. It covers 4.572 km by 4.572 km. A total of 91,031,226 observation points (4.35 points/$m^2$) were acquired from an airborne laser scanning. 
The RGB aerial image and the finely rasterized DSM were shown in Figure \ref{fig:fig4}. For our DTM generation method, all parameters were selected as default values described in Section 2. For LAS, noise removal preceded as recommended in the Lastools documentation, and default parameters were used for DTM generation. For CSF, the “relief” scenario of CloudCompare \footnote{https://www.cloudcompare.org/} plug-in was adopted, and other settings were set as a default. All DTMs were generated with 0.5-meter resolution.

To effectively evaluate in a large area, we adopted a tiling comparison method. The tiling comparison is a method that compares maps by dividing them into small tiles \citep{song2022towards}. Conventionally, ground filtering methods have compared their performances by regarding it as a classification task that determines whether a given point is ground or non-ground \citep{hu2015semi,mongus2013computationally,sithole2004experimental}. Under the premise that there is a high quality of ground truth points, this method can provide clear comparative, quantitative results among different ground filtering algorithms as the ground filtering algorithm itself is to classify ground and non-ground points. However, it has a limitation in that accurate ground truth is hard to be obtained for a large area, resulting in limited experimental areas \citep{meng2010ground,polidori2020digital}. Thus, we adopted the tiling comparison method to effectively compare DTM generation methods. Also, quantitative measurements, the mean absolute error (MAE) and the root mean square error (RMSE) are also provided for the comparison as other image grid-based studies adopted \citep{chen2012upward,hu2016deep,gevaert2018deep,amirkolaee2022dtm}.

To be specific, we tiled the entire DTM of the Purdue University dataset into 81 tiles so that area of each tile is to be 0.5 km by 0.5 km. Then, we ranked the DTMs based on MAE between our DTM and others. The MAE was calculated by comparing all pixel elevation values of our DTM to other two DTMs, respectively. Likewise, RMSE between our DTM and others was also computed. As water elevations of LAS and CSF were either 0 or significantly lower values, which are not reliable, we masked the water area before computing MAE and RMSE.

\subsection{Experimental results}
\label{sec:Experimental results}

This subsection provides the comparison results among OUR, CSF, and LAS. We excerpted four tiles that show distinctive differences among different methods and that can provide helpful information to potential users. 

\begin{figure}
	\centering
	\includegraphics[width=1\textwidth]{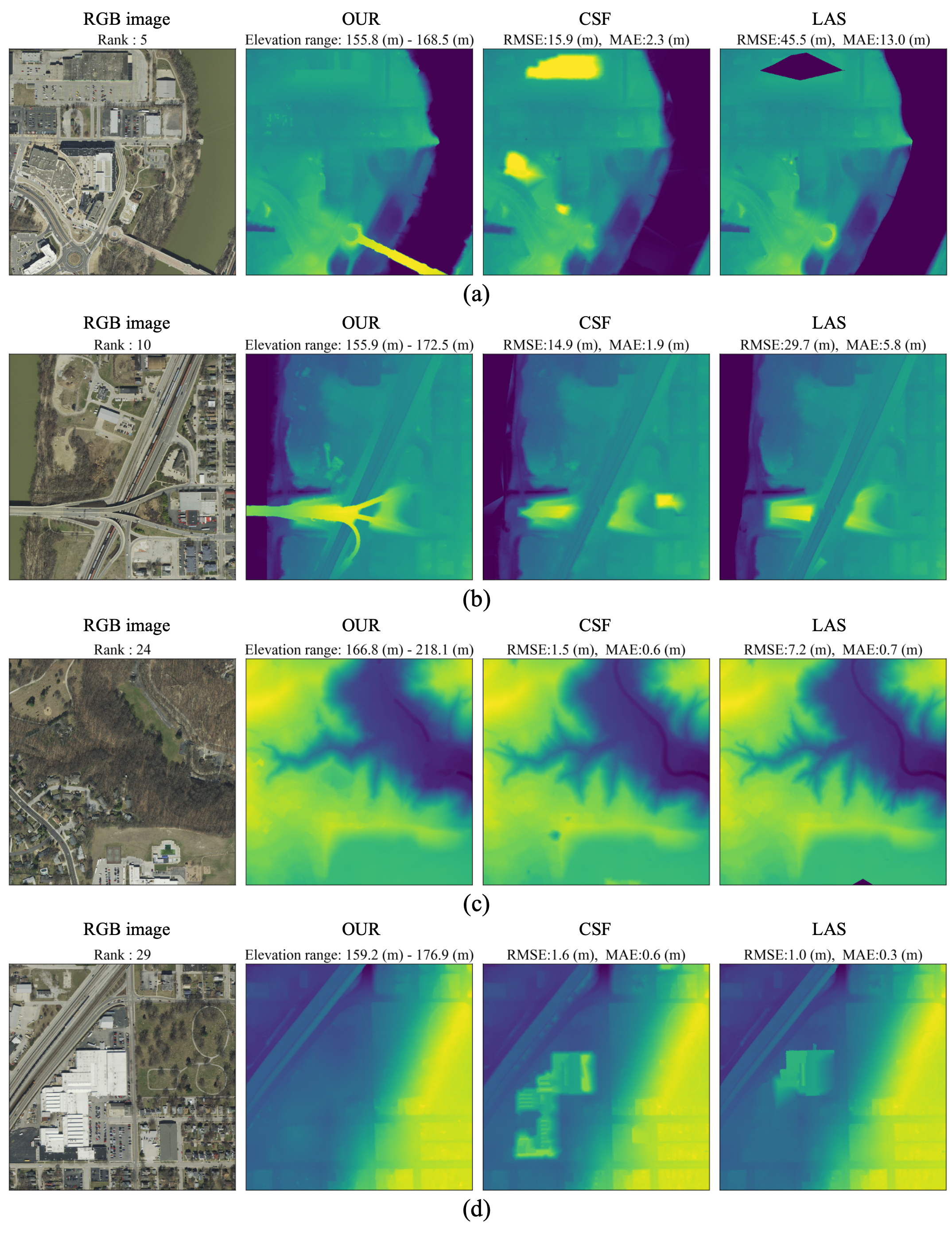}
	\caption{Comparison of DTMs generated from different methods. 4 out of 81 tiles that showed significant and distinctive difference among different DTMs were excerpted}
	\label{fig:fig5}
\end{figure}

Figure \ref{fig:fig5} shows aerial RGB images and three different DTMs from different methods. The RGB images are from the U.S. Department of Agriculture’s (USDA) National Agriculture Imagery Program (NAIP)'s orthoimagery. The rank denoted with RGB images indicates the order of the highest MAE values out of 81 tiles. The elevation ranges of OUR's DTM were provided for reference. The RMSE and MAE values of other DTMs calculated by comparing to OUR's DTM were also provided.

Figure \ref{fig:fig5} (a) displays an urban area along the river. CSF and LAS were not able to generate proper DTM for the large building. CSF regarded the large building as the ground while LAS produced a hole (0 value) as same as the river. On the other hand, OUR filtered out the building as a non-ground and interpolated it with nearby ground elevations. Another unique difference can be found in the bridge. As the bridge is connected to the ground without a discrete elevation change, OUR classified the bridge into a ground category. However, neither CSF nor LAS considered the bridge as a ground object. 

Figure \ref{fig:fig5} (b) shows the overpass structures. As the overpass is connected to the ground smoothly, OUR classified the overpass as ground. However, both CSF and LAS considered the overpass as non-ground. This is because CSF simulates a cloth covering on the upside-down DSM. The TIN-based method, LAS, also produced a similar result. Another difference is that OUR regarded the pile of soil as the ground while CSF and LAS removed it from the ground category.

Figure \ref{fig:fig5} (c) illustrates the disadvantage of OUR as it overly smoothed deep valleys (i.e., West Lafayette Parks Maintenance). The reason why the small, narrow valleys were smoothed is that some narrow valleys were enclosed by steep terrains. 

Figure \ref{fig:fig5} (d) shows the advantage of OUR in that our method was able to filter out a large building as a non-ground and produced a reliable DTM while a significant portion of the large building remains in both DTMs of CSF and LAS. 

In summary, OUR produced more reliable DTMs compared to CSF and LAS, particularly for urban areas with large buildings. Also, OUR has a unique characteristic that can map bridges and overpasses as a category of the ground. It does not mean OUR falsely classified bridges and overpasses just because bridges and overpasses are human-made structures. Although it is not a natural terrain, it is an artificial terrain like roads and plays a role more like a ground. Moreover, the terrain under the bridge and overpasses are actually unknown. In addition, note that both CSF and LAS partially mapped overpasses as ground as shown in Figure \ref{fig:fig5} (b). CSF and LAS simply interpolated unmeasured grounds after removing the layer on top. Also, considering CSF and LAS classified some parts of the overpass as ground classes, we can argue that CSF and LAS were not consistent in their decision rules for overpasses, and their results are difficult to predict and explain. Rather than debating whether bridges and overpasses are ground or not, it is worth noting that each DTM definition has its own merits. DTM that regarding bridges and overpasses as the terrain can be beneficial in some applications where overpasses and bridges should be considered as ground such as building extraction \citep{song2022towards}. Lastly, OUR method with the default parameter overly smoothed some steep, narrow areas as shown in Figure \ref{fig:fig5} (c). However, it can be prevented by tuning the slope threshold. The impact of the slope threshold is detailed in Section 3.2.

\subsection{Suggestions for parameter selection}
\label{sec:Suggestions for parameter selection}

\subsubsection{Slope threshold}
\label{sec:Slope threshold}

The slope threshold is the most important parameter because our method is based on the unique assumption that non-ground objects are surrounded by break-lines, and break-lines are determined by the slope threshold. We set the slope threshold to 45 degrees as default in experiment 3.1. and found the generated DTM is reliable where terrain relief is moderate (approximately lower than 45 degrees). However, DTM tends to get blurred where the areas are surrounded by steep slopes like a deep narrow valley. 

\begin{figure}
	\centering
	\includegraphics[width=1\textwidth]{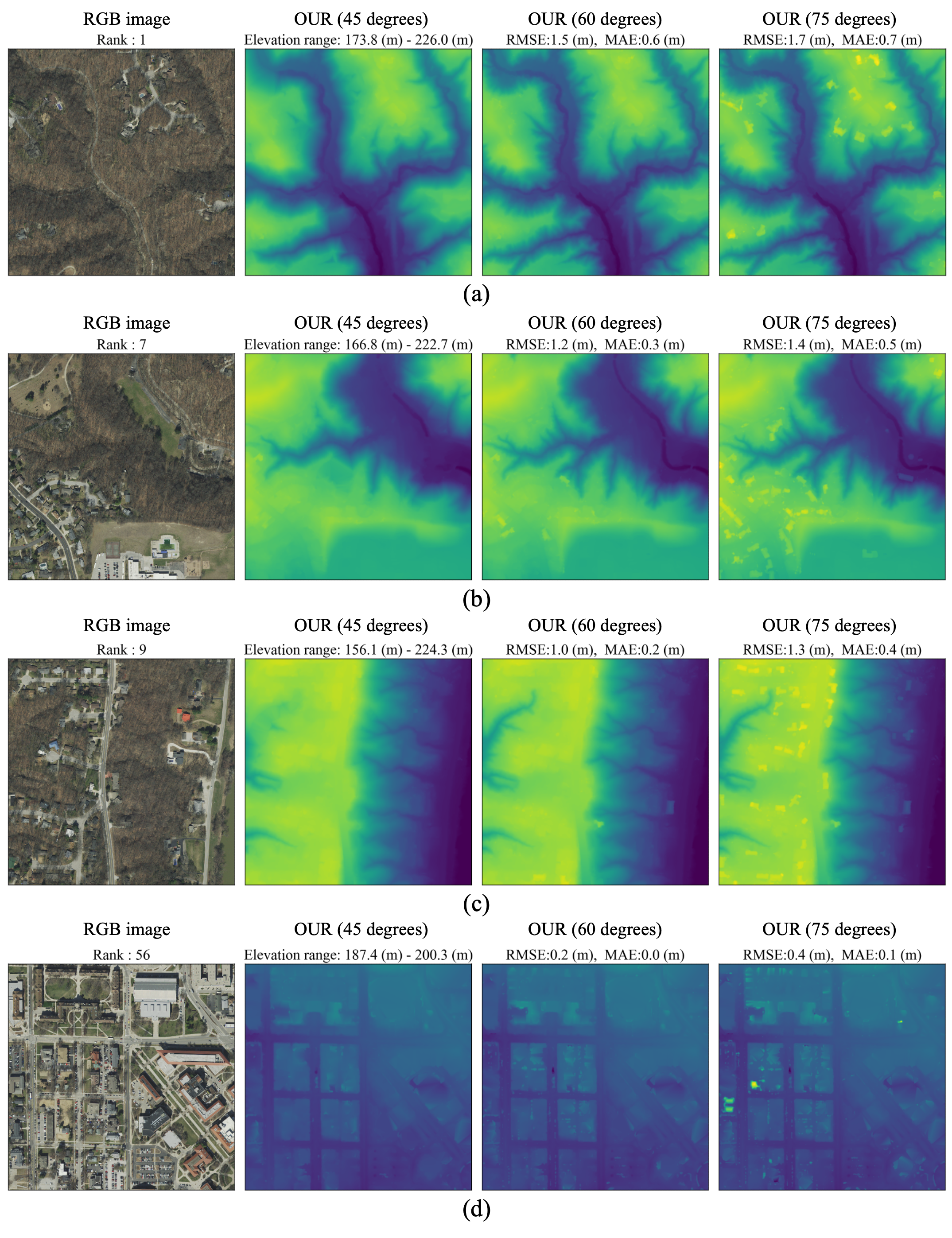}
	\caption{Comparison of DTMs generated from OUR with different slope thresholds}
	\label{fig:fig6}
\end{figure}

To investigate the impact of the slope threshold, we investigated the DTMs when the slope thresholds were set to different values. A total of three slope thresholds (i.e., 45 degrees, 60 degrees, and 75 degrees) were selected, and their resultant DTMs are shown in Figure \ref{fig:fig6}. The rank indicates the order of the highest MAE values compared to results of 45 degrees. DTM elevation ranges are provided. MAE and RMSE values that are compared to DTM with the default slope threshold (45 degrees) are provided as well.

As a result, it was found that the impact of the slope threshold was distinct in mountainous and hilly areas as shown in Figure \ref{fig:fig6} (a-c). Compared to the DTM with a default slope threshold (45 degrees), DTMs with higher slope thresholds were able to delineate reliable terrain maps near steep and narrow valleys. However, several buildings particularly near hilly areas were identified as terrain. This is because some buildings could have been connected to the terrain with less than 60 or 75 slope degrees. In the flat urban area, however, buildings were identified as a terrain in most cases. 

\subsubsection{A1, A2, and R}
\label{sec:A1, A2, and R}

A1, A2, and R are to prevent the case where the ground is misclassified as non-ground when the disconnected, small area of ground lies on the edge of the LiDAR file data extent (Please refer to Figure \ref{fig:fig3}). Therefore, only A2 will be required if a wider range of point clouds surrounding the target area are available. For example, in practical usage, users can generate DTM by gathering larger surrounding coverage of point clouds than the target of interest, and crop the centered target area to prevent errors. However, it will increase a computational cost and there must be a case when the larger data extent is not available. To address this practical issue, A1, A2, and R were used. The following illustration shows the error near the edge of the data extent and how those parameters can resolve the error. 

\begin{figure}
	\centering
	\includegraphics[width=1\textwidth]{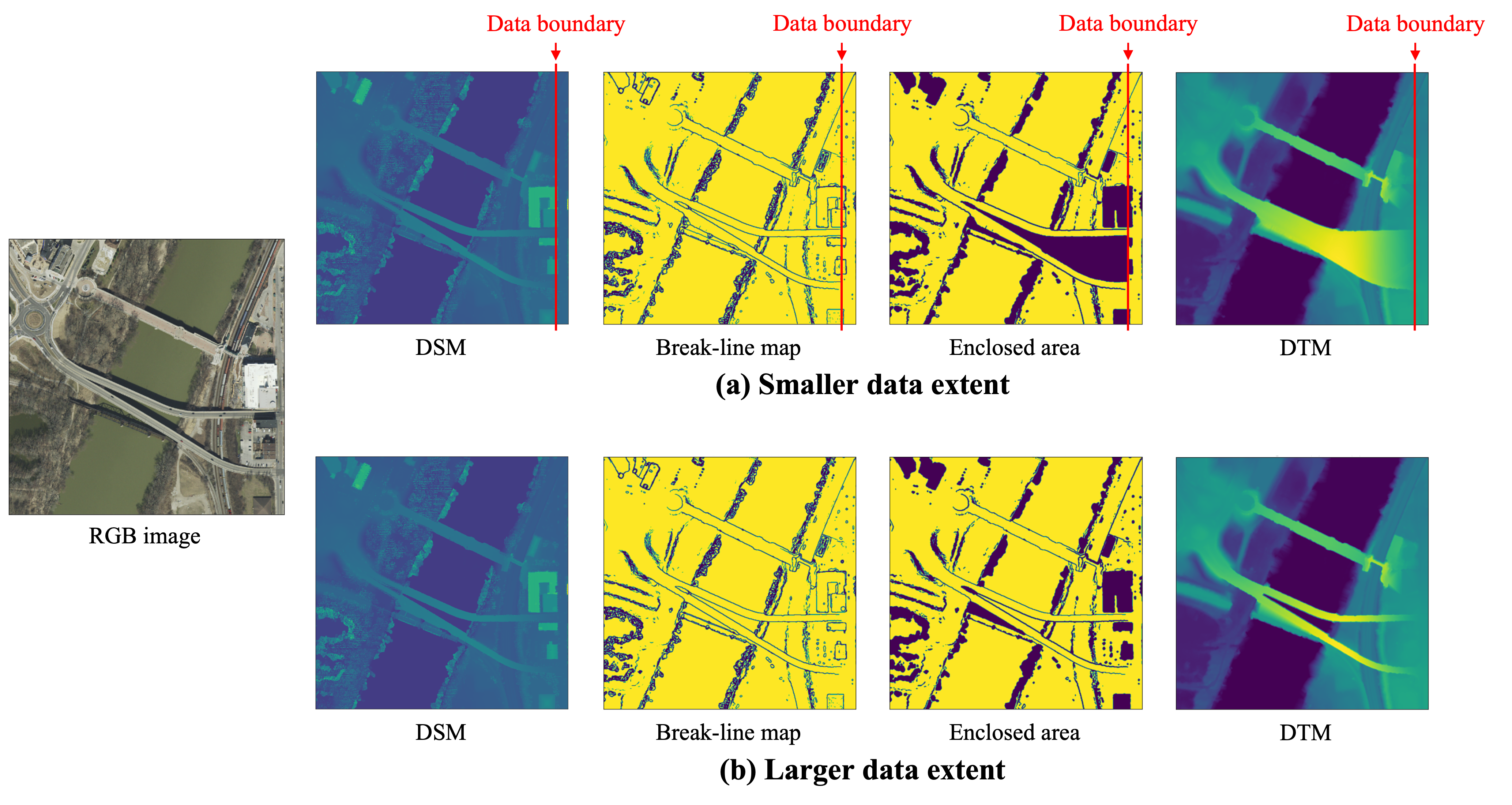}
	\caption{An example of an error near the data boundary and a remedy for the error}
	\label{fig:fig7}
\end{figure}

Figure \ref{fig:fig7} shows an example of an error near the data boundary and a remedy for the error. Figure \ref{fig:fig7}(a) (“smaller data extent”) is the case when the LiDAR files were divided and processed separately while Figure \ref{fig:fig7}(b) (“larger data extent”) shows the case when the LiDAR files were merged and processed together. In the smaller data extent, the diverging bridge ends with the tile boundary and ends up enclosing the interim area between diverging bridges. Since the interim area was smaller than A1, the area was classified as a non-ground. If this area was larger than the A1, depending on parameters of A2 and R, the area could have been determined as a ground. In this example, the default parameter was not able to classify the interim area properly. However, when the larger data extent is used, the interim area is eventually connected to a larger ground area, and eventually, the combined area of the interim area and larger ground areas must have been either larger than A2 or larger than A1 and smaller than R. In the end, the larger data extent produced a reliable DTM.

\subsubsection{Water-related parameters}
\label{sec:Water-related parameters}

Water is identified by considering the distribution of the number of points in the given window. Normal distribution was assumed to find water pixels considering that the point density over the water area is significantly lower than that of the non-water area. As all parameters associated with water body mapping were parameters for the normal distribution, the parameter setting could be easy and intuitive. Unless the point distribution of the scan is severely imbalanced or the laser scan contains a large occluded area (e.g., near tall buildings), the default parameter would perform well with most of the topographic near-infrared LiDAR data.

\begin{figure}
	\centering
	\includegraphics[width=1\textwidth]{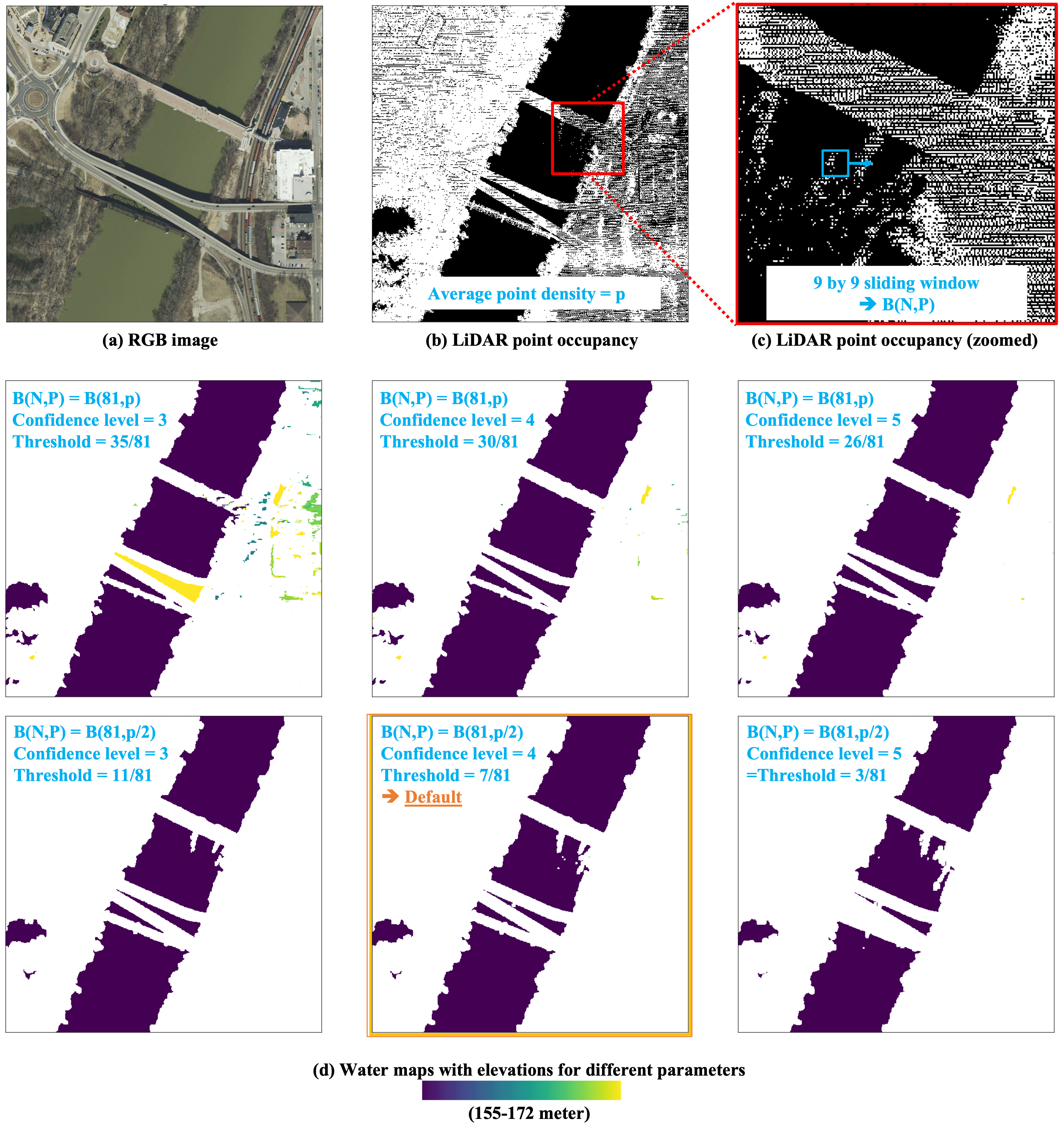}
	\caption{Water mappings and the impacts of water-related parameters}
	\label{fig:fig8}
\end{figure}

Figure \ref{fig:fig8} illustrates a water mapping and the impacts of water-related parameters. Figure \ref{fig:fig8}(a) displays a RGB image for reference. Figure \ref{fig:fig8}(b) shows LiDAR point occupancy that shows the grid occupied by the LiDAR points in white and otherwise in black. Figure \ref{fig:fig8}(c) shows a zoomed-in image of Figure \ref{fig:fig8}(b). Figure \ref{fig:fig8}(d) illustrates the results of water mapping for different parameter combinations. As described in Section 2.4., we used B(N, P/2) and a confidence level of 4 as a default, where N is the number of pixels in a sliding window and P is the average point density. The size of the window was 9 by 9 (81 pixels), and the average point density was 0.6. As a result, the central pixel of the window whose number of LiDAR points is less than the threshold (7) out of 81 was classified as water. As the threshold decreases, water segments tend to be less detected and smaller. After water segments were mapped, the 10th percentile of elevations among each water segment was used for the elevation of the segment. 

\subsection{Limitations}
\label{sec:Limitations}

Since DTM generation involves the binary classification between non-ground and ground, our method shares the common limitation of binary classification that has a trade-off between omission and commission errors. To be specific, as the slope threshold increases, the DTM of the steep area retains sharper terrain reliefs, but some non-ground objects can remain as ground. Conversely, if the slope threshold decreases, the DTM is getting more smoothed while it could prevent the presence of non-ground objects in the resulting DTM. This limitation commonly exists in DTM generation algorithms and the trade-off can be controlled by several parameter tunings in some algorithms \citep{chen2017state,liu2008airborne,meng2010ground}. For example, CSF can tune the rigidness of cloth \citep{zhang2016easy}. LAS can tune the parameter for the maximal standard deviation for planar patches of TIN \citep{axelsson2000generation}. It is worth mentioning that the parameter setting of our method is very straightforward and intuitive, and thus, the outcome according to the parameter is easily predictable, compared to other methods. This advantage enables users to find the proper parameters for their objectives and study areas and can significantly reduce uncertainties in the resulting DTM.

Although the proposed algorithm was confirmed to be robust to generate DTM of diverse topography, our algorithm can suffer when only the physical shape of the target is not enough to identify whether it is a ground or non-ground. This limitation is common in most DTM generation algorithms. For example, there is no way to distinguish between dome-shaped buildings and the same shape of rocky terrain unless a sophisticated semantic understanding is possible. Adding more delicate decision rules might enable the algorithm to distinguish some rare but difficult cases, but it will compromise the algorithm’s generalization capability and computational efficiency. Deep learning-based algorithms that can make a decision based on the learned distribution may work better, particularly when sophisticated semantic understanding is needed. However, it will necessarily entail errors from the distribution shift where the distribution of the target area is different from that of the trained area \citep{moreno2012unifying,tuia2016domain}. Also, another limitation of deep learning-based algorithms is their decisions are necessarily limited by their input size \citep{amirkolaee2022dtm,gevaert2018deep,hu2016deep}. The typical input size of deep learning-based models for semantic segmentation is 256 by 256. Therefore, the model should decide whether the target is ground or non-ground based on the limited area of 256-meter by 256-meter if the resolution of DSM is 1-meter. However, if the input is composed of only the center of a large flat building, the deep model will be likely to fail in its mapping. Errors in large object detection frequently occur in deep learning-based methods \citep{song2022towards,ji2018fully}. On the other hand, our proposed algorithm can consider the entire data extent for the decision. In other words, inputs do not need to be tiled like in deep learning. More importantly, due to uncertainties of the decision rule of deep learning methods, the so-called “black box”, predicting and explaining the results would be difficult, which could jeopardize the credibility of the subsequent analyses. Of course, our algorithm is not without errors. However, the magnitude and influence of the error can be better estimated than other algorithms as parameter tunings are very intuitive and the result of the algorithm can be easily explained.

Another limitation to be noted is the uncertainties in water elevation and water mapping. Our DTM mapping workflow includes a feature for water body mapping to alleviate elevation errors near water bodies and to assist subsequent studies. Having the water mapping in our workflow is advantageous as users can expect a DTM having all elevations in both terrain and water, and it can replace the post-processing of water mapping that requires another external source of data. However, due to the low reflectance of a near-infrared laser to water, observations include lots of noise, resulting in uncertainties of water elevation. In fact, even if the MAE was calculated after masking the water area when performing the tile comparison in Figure \ref{fig:fig5}, tiles containing large water bodies recorded the highest MAE (18 out of the top 20 contain water bodies either river or lake). This suggests that the water mask was not perfect and that the CSF and LAS also had errors near water bodies. Particularly, we witnessed water with fast flowing streams and lots of floating often has high point density, resulting in being omitted from the water mask. There have been previous studies that map water with airborne LiDAR data by using a supervised classification \citep{brzank2008aspects,smeeckaert2013large} or LiDAR signal intensity \citep{hofle2009water}. Although they require either training procedures \citep{brzank2008aspects,smeeckaert2013large} or a radiometric correction of intensity \citep{hofle2009water}, they could be an alternative way of mapping water bodies. Future studies for more accurate and scalable water mapping with airborne LiDAR data are needed. 

\section{Conclusion}
\label{sec:Conclusion}

We developed an open-source algorithm for DTM generation using airborne LiDAR data. The proposed DTM generation algorithm is based on a novel assumption that every non-ground object is surrounded by a certain level of a steep slope while grounds are smoothly connected to each other. With the tiling evaluation, we compared distinctive differences among different algorithms effectively and confirmed that our algorithm can produce reliable DTM in diverse landscapes. Lots of DTM generation algorithms have been proposed for decades. Most algorithms process the ground filtering based on the point clouds and their association to the local neighborhood, which can be susceptible to varying local point densities and data noise. On the other hand, our method does not require a concept of either local neighborhood or search radius, which makes the algorithm robust and think beyond the local neighborhood. Also, as our algorithm works based on a simple but robust assumption, it involves few parameters, and parameters can be tuned very intuitively. Moreover, our algorithm includes a water mapping feature and can produce DTM whose errors are more manageable and predictable, compared to deep learning-based DTM extraction methods. But, it is worth to mention the algorithm may require different parameters for mountainous areas and flat terrains, respectively. Future works would develop a way to automatically tune the parameter based on the local landscape. Our algorithm and data used in the experiment will be open to the public.



\bibliographystyle{elsarticle-harv} 
\bibliography{main}





\end{document}